\def\year{2021}\relax
\documentclass[letterpaper]{article} 
\usepackage{aaai21}  
\usepackage{times}  
\usepackage{helvet} 
\usepackage{courier}  
\usepackage[hyphens]{url}  
\usepackage{graphicx} 
\usepackage{graphicx}  
\usepackage{natbib}  
\def\x{\mathbf{x}}
\usepackage{placeins}
 
\usepackage{caption} 
 \usepackage{algorithm}
\usepackage{algpseudocode}
\usepackage{tabularx}
 \usepackage{verbatim}
\usepackage{bm}
\usepackage{lipsum}
\usepackage{tabularx}
\usepackage{mwe} 
\usepackage[utf8]{inputenc} 
\usepackage{url}            
\usepackage{booktabs}       
\usepackage{nicefrac}       
\usepackage{microtype}      
\usepackage{times}
\usepackage{epsfig}
\usepackage{graphicx}
\usepackage{amsmath}
\usepackage{amsthm}
\usepackage{mathrsfs} 
\usepackage{amssymb}
\usepackage{wrapfig}
\usepackage{lscape}
\usepackage{rotating}
\usepackage{subfig}
\usepackage[switch]{lineno}
\usepackage{adjustbox}

\usepackage{wrapfig}

\usepackage{bm}

\usepackage{setspace}
\usepackage{xspace}

\usepackage{breakcites}

\usepackage{xcolor}

\newcommand*\rot{\rotatebox{90}}
\usepackage{blindtext}
\title{Unsupervised Model Adaptation for Continual Semantic Segmentation}

\author{
    \normalfont 
    \begin{tabular}{*{2}{>{\centering}p{.5\textwidth}}}
    Serban Stan & Mohammad Rostami \tabularnewline
    University of Southern California & University of Southern California \\ Information Sciences Institute \tabularnewline
    \url{sstan@usc.edu} & \url{mrostami@isi.edu} 
    \end{tabular}
}

\begin{document}

\maketitle

\begin{abstract}
We develop an algorithm for adapting a semantic segmentation model that is trained using a labeled source domain to generalize well in an unlabeled target domain. A similar problem has been studied extensively in the unsupervised domain adaptation (UDA) literature, but existing UDA algorithms require access to both the source domain labeled data and the target domain unlabeled data for training a domain agnostic semantic segmentation model.  Relaxing this constraint enables a user to adapt  pretrained models to generalize in a target domain, without requiring access to source data. To this end, we learn a prototypical distribution for the source domain in an intermediate embedding space. This distribution encodes the abstract knowledge that is learned from the source domain. We then use this distribution for aligning the target domain distribution with the source domain distribution in the embedding space. We provide theoretical analysis and explain conditions under which our algorithm is effective. Experiments on  benchmark adaptation task demonstrate our method achieves competitive performance even compared with joint UDA approaches. 
\end{abstract}

\section{Introduction}
Image segmentation is an essential computer vision ability for  delivering technologies such as autonomous driving~\cite{zhang2016instance} and automatic object tracking~\cite{wang2019fast}. Advances in deep learning have led to the development of image segmentation algorithms with close to human-level performance~\cite{zeng2017deepem3d}. However, this success is conditioned on the availability of large and high-quality manually annotated datasets to satisfy the required sample complexity bounds for training generalizable deep neural networks.  As a result, data annotation is a major bottleneck to address   the problem of \textit{domain shift}, where  \textit{domain discrepancy} exists between the distributions of training and testing  domains~\cite{luo2019taking} and the trained model needs to be adapted to generalize again after being fielded. This is particularly important in continual learning~\cite{shin2017continual}, where the goal is to enable a learning agent to learn new domains autonomously. Retraining the model from scratch is not a feasible solution for continual learning because manual data annotation is an expensive and time-consuming process for image segmentation, e.g., as much as 1.5 hours for a single image of the current benchmark datasets~\cite{cordts2016cityscapes}. A practical alternative is to adapt the trained model using only unannotated data. 

The problem of model adaptation for image segmentation has been studied extensively in the unsupervised domain adaptation (UDA) framework. The goal in UDA is to train a model for an unannotated target domain by transferring knowledge from a secondary related source domain in which annotated data is accessible or easier to generate, e.g., a synthetically generated domain. Knowledge transfer can be achieved by extracting domain-invariant  features from the source and the target domains to address domain discrepancy. As a result, if we train a classifier using the source domain   features as its input, the classifier will generalize on the target domain   since the distributions of features are indistinguishable. Distributional alignment can be achieved by matching the distributions at different levels of abstraction, including appearance~\cite{hoffman2017cycada,sankaranarayanan2018learning}, feature~\cite{hoffman2017cycada,murez2018image},  output~\cite{zhang2017curriculum,rostami2019learning} levels.

A large group of the existing UDA algorithms for image segmentation use adversarial learning for extracting domain-invariant features~\cite{luc2016semantic,bousmalis2017unsupervised,hoffman2018cycada,murez2018image,saito2018maximum,sankaranarayanan2018learning,dhouib2020margin}. Broadly speaking, a domain discriminator network can be trained to distinguish whether an input data point comes from the source or the target domain. This network is fooled by a feature generator network which is trained to make the domains similar at its output. Adversarial training~\cite{goodfellow2014generative}  of these two networks leads to learning a domain-agnostic embedding space.  A second class of UDA algorithms   directly minimize suitable loss functions that enforce domain alignment~\cite{chen2017importance,wu2018dcan,zhang2017curriculum,zhang2019category,lee2019sliced,yang2020fda}.
Adversarial learning requires delicate   optimization initialization, architecture engineering, and careful selection of hyper-parameters to be stable~\cite{roth2017stabilizing}. In contrast, defining a suitable loss function for direct domain alignment may not be trivial.

A major limitation of existing UDA algorithms is that domain alignment can be performed only if the source and the target domain data are accessible concurrently. However, the source annotated data may not be necessarily accessible during the model adaptation phase in a continual learning scenario. In this paper, we focus on a more challenging, yet more practical model adaptation scenario. We consider that a pretrained model is given and the goal is to adapt this model to generalize well in a target domain using solely unannotated target domain data.   Our algorithm can be considered as an improvement over using an off-the-shelf   pre-trained model naively by benefiting from the unannotated data in the target domain. This is a step towards lifelong learning ability~\cite{shin2017continual,rostami2019complementary}. 

{\em \bf Contributions:} our main  contribution is to relax the need for source domain annotated data for model adaptation. Our idea is to learn a   prototypical distribution that encodes the abstract knowledge, learned for image segmentation using the source domain annotated data.  The prototypical distribution is used for aligning the distributions across the two domains in an embedding space.   We also provide  theoretical analysis to justify the proposed model adaptation algorithm and determine the conditions under which our algorithm is effective. Finally, we provide experiments on the GTA5$\rightarrow$Cityscapes and SYNTHIA$\rightarrow$Cityscapes benchmark domain adaptation image segmentation tasks to demonstrate that our method is effective and leads to competitive performance, even when compared against existing UDA algorithms.

\section{Related Work}

 We discuss related work on domain adaptation and semantic segmentation, focusing on direct distribution alignment.

 \subsection{Semantic Segmentation}
 Traditional semantic segmentation algorithms use hand-engineered extracted features which are fed into a classifier~\cite{shotton2008semantic,tighe2010superparsing}, where the classifier is trained using supervised learning. In contrast, current state of the art approaches use convolutional neural networks (CNNs) for feature extraction~\cite{long2015fully}. A base CNN subnetwork is converted into a fully-convolutional network (FCN) for feature extraction and then is combined with a classifier subnetwork to form an end-to-end classifier. The whole pipeline is trained in an end-to-end deep supervised learning training scheme.  Due to large size of learnable parameters of the resulting semantic segmentation network, a huge pixel-level manually annotated dataset is required for training.

 We can use weakly supervised annotation such as using bounding boxes to reduce manual annotation cost~\cite{pathak2015constrained,papandreou2015weakly}, but even obtaining weakly annotated data for semantic segmentation can be time-consuming. Additionally, a trained model using weakly annotated datasets may not generalize well during testing.
  Another approach for relaxing the need for manually annotated datasets is to use synthetic datasets which are generated using computer graphics~\cite{ros2016synthia,cordts2016cityscapes}. These datasets can be annotated automatically. But a trained model using synthetic datasets, might not generalize well to real-world data due to the existence of domain shift problem~\cite{sankaranarayanan2018learning}. 
   Unsupervised domain adaptation is developed to address this problem. 

 \subsection{Domain Adaptation}
 Domain adaptation methods reduce domain discrepancy by aligning distributions using annotated data in a target domain and unannotated data in a target domain. A group of UDA methods use a shared cross-domain encoder to map data into a shared embedding space and train the encoder by  minimizing a  probability distance measure across the two domains at its output. The Wasserstein distance (WD)~\cite{courty2017optimal,damodaran2018deepjdot} is an example of such measures which captures higher-order statistics.
 Damodaran et al.~\cite{damodaran2018deepjdot} demonstrate that using WD leads to performance improvement over methods  that rely on matching lower-order statistics~\cite{long2015learning,sun2016deep}.
In this work, we   rely on  the sliced Wasserstein distance (SWD) variant of WD~\cite{lee2019sliced} for domain alignment~\cite{rostami2019deep}. SWD has a closed form solution and can be computed more efficiently than WD.

Current UDA methods assume that the source and the target domain data are accessible concurrently during domain alignment. However, since usually a model is already pre-trained on the source domain, it is beneficial if we can  adapt it using the target domain unannotated data. This model adaptation setting been explored for non-deep models~\cite{dredze2008online,jain2011online,wu2016online}, but these works cannot be extended  to semantic segmentation tasks. In this work, we benefit from prototypical distributions to align two distributions indirectly.  The core   idea is that  the image pixels that belong to each semantic class form a   data cluster in a shared embedding space.  The centroid for this cluster is called class prototype. Recently, UDA has been addressed by aligning the prototype pairs across two domains~\cite{pan2019transferrable,chen2019progressive,rostami2020sequential}. Inspired by these works, we  extend the notion of class prototypes to prototypical distributions  in the embedding space. A prototypical distribution for image segmentation is a multimodal distribution that encodes the knowledge learned from the source domain. 
Our work is based on enforcing the two domains to share a similar prototypical distribution in the embedding.

\section{Problem Formulation}

\begin{figure*}[!htb]
    \centering
     \includegraphics[width=.8\textwidth]{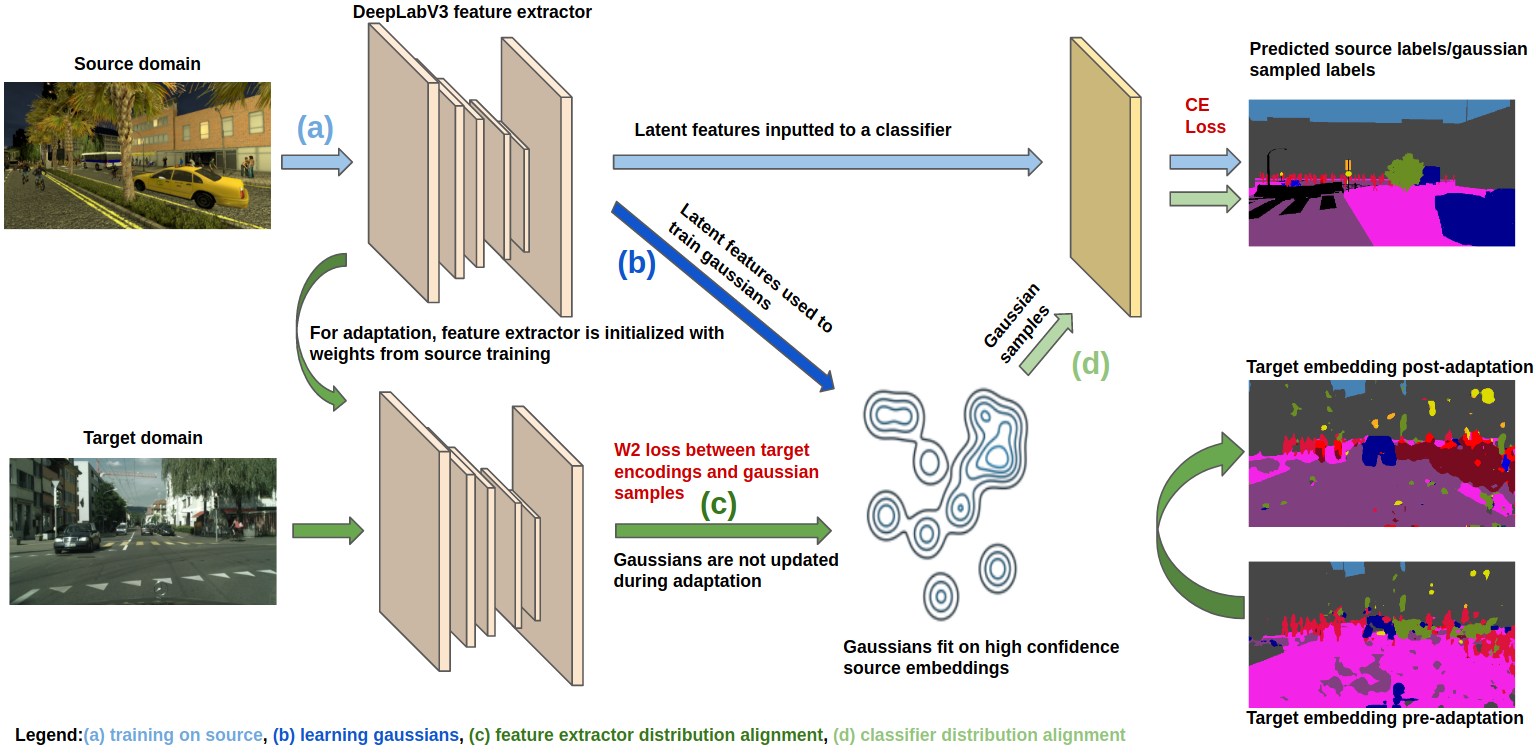}
    \caption{Diagram of the proposed model adaptation approach (best seen in color): (a) initial model training using the source  domain labeled data, (b) estimating the prototypical distribution as a GMM distribution in the embedding space, (c) domain alignment is enforced by minimizing the distance between the prototypical distribution samples and the target unlabeled samples, (d) domain adaptation is enforced for the classifier module to fit correspondingly to the GMM distribution.}
     \label{ISfig:blockdiagram}
\end{figure*}

Consider an image domain $\bm{X}^{s}$ and  a semantic segmentation model $f_{\bm{\theta}}(\cdot):\bm{X}^{s}\rightarrow\bm{Y}^{s}$ with learnable parameters ${\bm{\theta}}$ which receives an input image $\bm{x}^{s}\in \bm{X}^{s}$ and predicts  pixel-wise category labels $\bm{y}^{s}\in\bm{Y}^{s}$. The goal is to train the model such that the expected error, i.e. true risk, between the prediction and the ground truth is minimized, i.e., $\bm{\theta}^* =\arg\min_{\bm{\theta}}\{\mathbb{E}_{\bm{x}^{s}\sim P_{\mathcal{S}}(\bm{X})^{s}}(\mathcal{L}(f_{\bm{\theta}}(\bm{x}^{s}),\bm{y}^{s})\}$, where $P_{\mathcal{S}}(\bm{X}^{s})$ and $\mathcal{L}(\cdot)$ denote the input data distribution and a suitable loss function, respectively. In practice, we use empirical risk minimization (ERM) and  the cross-entropy loss for  solving for the optimal semantic segmentation model:
\begin{equation}
\begin{split}
& \hat{\bm{\theta}} =\arg\min_{\bm{\theta}}\{\frac{1}{N} \sum_{i=1}^N\mathcal{L}_{ce}(f_{\bm{\theta}}(\bm{x}^{s}),\bm{y}^{s})\} 
\\&\mathcal{L}_{ce} =- \sum_{k=1}^K\sum_{h=1}^{H\times W} y_{ijk}\log(p_{ijk}),
\end{split}
\label{eq:MAPestIS}
\end{equation} 
where $N$ and $K$ denote the training dataset size and the number of semantic categories. $H$ and $W$  denote the input image height and width, respectively. Also, $\bm{y}_{ij} =[y_{ijk}]_{k=1}^K$ is a one-hot vector that denotes the ground-truth semantic labels and $\bm{p}_{ij} =[p_{ijk}]_{k=1}^K$ is a probability vector of  the predicted category probabilities by the model, i.e., a softmax layer is used as the last model layer. In practice, the model is fielded after training for testing and we do not store the source samples. If $N$ is large enough and the base model is complex enough, the ERM-trained model will generalize well on unseen samples, drawn from the distribution $P_{\mathcal{S}}(\bm{X}^{s})$. 

Now, consider that after  training, we want to  employ the source-trained model in a target domain $\bm{X}^{t}$ with the distribution $P_{\mathcal{T}}(\bm{X}^{t})$, where $P_{\mathcal{T}}(\bm{X}^{t})\neq P_{\mathcal{S}}(\bm{X}^{s})$. Note that this situation emerges naturally in continual learning~\cite{shin2017continual} when data distribution changes over time. Within domain adaptation learning setting, this means that the target domain is encountered sequentially after learning the source domain. Due to existing distributional discrepancy, the source-trained model $f_{\hat{\bm{\theta}}}$ will have poor generalization capability on the target domain. Our goal is to improve the model generalization by adapting the model such that the source and the target domains share a similar distribution in an embedding space. Following the-state-of-the-art semantic segmentation models~\cite{long2015fully}, we consider a  deep network as the base model $f_{\bm{\theta}}(\cdot)$. This network is decomposed    into a deep CNN encoder $\phi_{\bm{u}}(\cdot): \mathbb{R}^{H\times W}\rightarrow   \mathbb{R}^P$, a category-wise CNN decoder $\psi_{\bm{v}}(\cdot): \mathbb{R}^P\rightarrow \mathbb{R}^{H\times W \times K}$, and a pixel-level classifier subnetwork $h_{\bm{w}}(\cdot):  \mathcal{Z}\subset \mathbb{R}^K\rightarrow \mathbb{R}^K$ such that $f_{\bm{\theta}} = h_{\bm{w}}\circ  \psi_{\bm{v}}\circ \phi_{\bm{u}}$, where $\bm{\theta}=(\bm{w},\bm{v},\bm{u})$ 
In this decomposition,  $\mathcal{Z}$ denotes the shared cross-domain embedding space. If we adapt the base trained model $f_{\hat{\bm{\theta}}}$ such that  the domain discrepancy is minimized, i.e., the distance between the distributions $\psi(\phi(p_{\mathcal{S}}(\bm{X}^s))$ and $\psi(\phi(p_{\mathcal{T}}(\bm{X}^t))$ is minimized in the embedding space, then  the source-trained classifier $h_{\hat{\bm{w}}}$ will generalize   on both domains. Most UDA algorithms benefit from this approach to address annotated data scarcity in the target domain but all assume that the source samples are accessible for model adaptation. Since this  makes computing the distance between the distributions $\psi(\phi(p_{\mathcal{S}}(\bm{X}^s))$ and $\psi(\phi(p_{\mathcal{T}}(\bm{X}^t))$ feasible, solving UDA reduces to aligning these   distributions. Note, however, $\psi(\phi(p_{\mathcal{T}}(\bm{X}^t))$ cannot be computed directly in our learning framework due to absence of source samples and we need   to estimate this distribution.

\section{Proposed Algorithm}
Figure~\ref{ISfig:blockdiagram} presents a high-level visual description our approach. Our solution is based on  aligning the source and the target distributions via an intermediate prototypical distribution in the embedding space. Since the last layer of the classifier is a softmax layer, we can treat the classifier as a maximum \textit{a posteriori} (MAP) estimator. This composition implies that if after training, the model can generalize well in the target domain, it must transform the source input distribution into a multimodal distribution
$p_J(\bm{z})$  with $K$ separable components in the embedding space (see Figure~\ref{ISfig:blockdiagram} (a)).  Each mode of this distribution represents one of the $K$ semantic classes. This prototypical distribution emerges as a result of model training because the classes should become separable in the embedding space for a generalizable softmax classifier. Recently, this property have been used for UDA~\cite{pan2019transferrable,chen2019progressive}, where the means for distribution modes are considered as the class prototype. The idea for UDA is to align the domain-specific prototypes for each class   to enforce distributional alignment across the domains. Our idea is to adapt the trained model using the target unlabeled data such that in addition to the prototypes, the source-learned prototypical distribution does not change after adaptation. As a result, the classifier subnetwork will still generalize in the target domain because its input distribution has been consolidated.

We model the prototypical distribution $p_J(\bm{z})$ as a Gaussian mixture model (GMM) with $k$ components: 
\begin{equation}
p_J(\bm{z})=\sum_{j=1}^k \alpha_j
\mathcal{N}(\bm{z}|\bm{\mu}_j,\bm{\Sigma}_j),
\label{gaussian}
\end{equation}  
where $\alpha_j$ denote mixture weights, i.e., prior probability for each semantic class. For each component, $\bm{\mu}_j$ and $\bm{\Sigma}_j$ denote the mean and co-variance of the Gaussian (see Figure~\ref{ISfig:blockdiagram} (b)). 

The empirical version of the prototypical distribution is accessible  by the source domain samples $\{(\psi_{\bm{v}}(\phi_{\bm{v}}(\bm{x}_i^{s})),\bm{y}_i^{s})\}_{i=1}^{N}$ which we use for estimating the parameters. Note that since the labels are accessible in the source domain, we can estimate the parameters of each component independently via MAP estimation. Additionally, since $p_{ijk}$ denotes the confidence of the classifier for the estimated semantic label for a given pixel. Hence, we can choose a threshold $\tau$ and compute the parameters using samples for which $p_{ijk}>\tau$ to cancel the effect of misclassified  samples that would act as outliers.  Let $\bm{S}_j$ denote the support set for class $j$ in the training dataset for which $p_{ijk}>\tau$, i.e., $\bm{S}_j=\{(\bm{x}_i^s,\bm{y}_i^s)\in \mathcal{D}_{\mathcal{S}}|\arg\max\hat{\bm{y}}_i^s=j, p_{ijk}>\tau \}$. Then, the MAP estimates for the distribution parameters would be:  
\begin{equation}
\small
\begin{split}
&\hat{\alpha}_j = \frac{|\bm{S}_j|}{\sum_{j=1}^N|\bm{S}_j|},\hspace{2mm}\hat{\bm{\mu}}_j = \sum_{(\bm{x}_i^s,\bm{y}_i^s)\in \bm{S}_j}\frac{1}{|\bm{S}_j|}\psi_u(\phi_v(\bm{x}_i^s)),\\& \hat{\bm{\Sigma}}_j =\frac{1}{|\bm{S}_j|}\sum_{(\bm{x}_i^s,\bm{y}_i^s)\in \bm{S}_j}\big(\psi_u(\phi_v(\bm{x}_i^s))-\hat{\bm{\mu}}_j\big)^\top\big(\phi_v(\phi_v(\bm{x}_i^s))-\hat{\bm{\mu}}_j\big).
\end{split}
\label{eq:MAPestAAS}
\end{equation}  

We   take advantage of the prototypical distributional estimate in Eq.\eqref{eq:MAPestAAS} 
as a surrogate for the source domain distribution to align the source and the target domain distribution in the absence of the source samples. We can adapt the model such that the encoder transforms the target domain distribution into the prototypical distributional in the embedding space. We use the prototypical distributional estimate  and draw random samples to generate a  labeled pseudo-dataset: $\mathcal{D}_{\mathcal{P}}=(\textbf{Z}_{\mathcal{P}},\textbf{Y}_{\mathcal{P}})$, where      $\bm{Z}_{\mathcal{P}}=[\bm{z}_1^p,\ldots,\bm{z}_{N_p}^p]\in\mathbb{R}^{K\times N_p}$,   $\bm{Y}_{\mathcal{P}}=[\bm{y}^p_1,...,\bm{y}^p_{N_p}]\in \mathbb{R}^{K\times N_p}$, $\bm{z}_i^p\sim \hat{p}_J(\bm{z})$. To improve the quality of the pseudo-dataset, we  use   the classifier sub-network prediction on drawn samples $\bm{z}^p$ to select   samples with $h_{\bm{w}}(\bm{z}^p) > \tau$. 
After generating the pseudo-dataset, we   solve the following optimization problem to align the source and the target distributions indirectly in the embedding:
\begin{equation}
\begin{split}
\arg\min{\bm{u},\bm{v},\bm{w}}&\{\frac{1}{N_p} \sum_{i=1}^{N_p}\mathcal{L}_{ce}(h_{\bm{w}}(\bm{z}_i^{(p)}),\bm{y}_i^{(p)})+\\&+\lambda D\big(\psi_{\bm{v}}(\phi_{\bm{v}}(p_{\mathcal{T}}(\bm{X}_{\mathcal{T}}))),\hat{p}_{J}(\bm{Z}_{\mathcal{P}})\big)\},  
\end{split}
\label{eq:mainPrMatchSS}
\end{equation}  
where $D(\cdot,\cdot)$ denotes a probability distribution metric to enforce alignment of the target domain distribution with the prototypical distribution in embedding space and $\lambda$ is a trade-off parameter between the two terms (see Figure~\ref{ISfig:blockdiagram} (c)).

  \begin{algorithm}[!htb]
\caption{$\mathrm{MAS^3}\left (\lambda , \tau \right)$\label{SSUDAalgorithmSS}} 
 {\small
\begin{algorithmic}[1]
\State \textbf{Initial Training}: 
\State \hspace{2mm}\textbf{Input:} source domain dataset $\mathcal{D}_{\mathcal{S}}=(\bm{X}_{\mathcal{S}},  \bm{Y}_{\mathcal{S}})$,
\State \hspace{4mm}\textbf{Training on Source Domain:}
\State \hspace{4mm} $\hat{ \theta}_0=(\hat{\bm{w}}_0,\hat{\bm{v}}_0\hat{\bm{u}}_0) =\arg\min_{\theta}\sum_i \mathcal{L}(f_{\theta}(\bm{x}_i^s),\bm{y}_i^s)$
\State \hspace{2mm}  \textbf{Prototypical Distribution Estimation:}
\State \hspace{4mm} Use Eq.~\eqref{eq:MAPestAAS} and estimate $\alpha_j, \bm{\mu}_j,$ and $\Sigma_j$
\State \textbf{Model Adaptation}: 
\State \hspace{2mm} \textbf{Input:} target dataset $\mathcal{D}_{\mathcal{T}}=(\bm{X}_{\mathcal{T}})$
\State \hspace{2mm} \textbf{Pseudo-Dataset Generation:} 
\State \hspace{4mm} $\mathcal{D}_{\mathcal{P}}=(\textbf{Z}_{\mathcal{P}},\textbf{Y}_{\mathcal{P}})=$
\State \hspace{12mm} $([\bm{z}_1^p,\ldots,\bm{z}_N^p],[\bm{y}_1^p,\ldots,\bm{y}_N^p])$, where:
\State \hspace{16mm} $\bm{z}_i^p\sim \hat{p}_J(\bm{z}), 1\le i\le N_p$
\State \hspace{17mm}$\bm{y}_i^p=
\arg\max_j\{h_{\hat{\bm{w}}_0}(\bm{z}_i^p)\}$, $p_{ip}>\tau$
\For{$itr = 1,\ldots, ITR$ }
\State draw random batches from $\mathcal{D}_{\mathcal{T}}$ and $\mathcal{D}_{\mathcal{P}}$
\State Update the model by solving Eq.~\eqref{eq:mainPrMatchSS}
\EndFor
\end{algorithmic}}

\end{algorithm}

The first term in Eq.~\eqref{eq:mainPrMatchSS} is to update the classifier such that it keeps its generalization power on the prototypical distribution. The second  term  is a matching loss term used to update the model such that the target domain distribution is matched to the prototypical distribution in the embedding space. Given a suitable probability metric, Eq.~\eqref{eq:mainPrMatchSS} can be solved using standard deep learning optimization techniques.

The major remaining question is selecting a proper probability  metric to compute $D(\cdot,\cdot)$. Note that the original target distribution is not accessible and hence we should select a metric that can be used to compute the domain discrepancy  via the observed target domain data samples and  the drawn samples from the prototypical distribution. Additionally, the metric should be smooth and easy to compute to make it suitable for gradient-based optimization  that is normally  used to solve Eq.~\eqref{eq:mainPrMatchSS}. In this work, we use Sliced Wasserstein Distance (SWD)~\cite{rabin2011wasserstein}. Wasserstein Distance (WD) has been used successfully for domain alignment in the UDA literature~\cite{courty2014domain,courty2017joint,bhushan2018deepjdot,xu2020reliable,li2020enhanced}. SWD is a variant of WD that can be computed more efficiently~\cite{lee2019sliced}.
SWD benefits from the idea of slicing by projecting high-dimensional probability distributions into their marginal one-dimensional distributions.  Since one-dimensional WD  has a closed-form solution, WD between these  marginal distributions can be computed fast. SWD   approximates WD as  a summation of WD between a number of random one-dimensional projections:
{\small
\begin{equation}
D(\hat{p}_J,p_\mathcal{T})\approx \frac{1}{L}\sum_{l=1}^L \sum_{i=1}^M| \langle\gamma_l, \bm{z}_{p_l[i]}^{p}\rangle- \langle\gamma_l, \psi(\phi(\x_{t_l[i]}^{t}))\rangle|^2
\label{SWDSemantic}
\end{equation}}
where $\gamma_l\in\mathbb{S}^{f-1}$ is uniformly drawn random sample from the unit $f$-dimensional ball $\mathbb{S}^{f-1}$, and  $p_l[i]$ and $t_l[i]$ are the sorted indices  for the prototypical and the target domain samples, respectively. We utilize Eq.~\eqref{SWDSemantic} to solve Eq.~\eqref{eq:mainPrMatchSS}.

 Our solution for source-free model adaptation, named  Model Adaptation for  Source-Free Semantic Segmentation (MAS$^3$), is described conceptually in Figure~\ref{ISfig:blockdiagram} and the corresponding algorithmic solution is given in  Algorithm~\ref{SSUDAalgorithmSS}.

 \section{Theoretical Analysis}
We analyze our algorithm within standard PAC-learning and prove that Algorithm~\ref{SSUDAalgorithmSS} optimizes an upper-bound of the expected error for the target domain under certain conditions.

Consider that  the hypothesis space within  PAC-learning is the set of  classifier sub-networks $\mathcal{H} = \{h_{\bm{w}}(\cdot)|h_{\bm{w}}(\cdot):\mathcal{Z}\rightarrow \mathbb{R}^k, \bm{w}\in \mathbb{R}^W\}$. Let $e_{\mathcal{S}}$ and  $e_{\mathcal{T}}$ denote the true expected error of the optimal domain-specific model from this space on the source and target domain respectively. We denote the joint-optimal model with $h_{\bm{w}^*}$. 
This model has the minimal combined source and target expected error $e_{\mathcal{C}}(\bm{w}^*)$, i.e. $\bm{w}^*= \arg\min_{\bm{w}} e_{\mathcal{C}}(\bm{w})=\arg\min_{\bm{w}}\{ e_{\mathcal{S}}+  e_{\mathcal{T}}\}$. In other words, it is a model with the best performance for both domains.

Since we process the observed data points from these domains, let  $\hat{\mu}_{\mathcal{S}}=\frac{1}{N}\sum_{n=1}^N\delta(\psi(\phi_{\bm{v}}(\bm{x}_n^s)))$ and $\hat{\mu}_{\mathcal{T}}=\frac{1}{M}\sum_{m=1}^M\delta(\psi(\phi_{\bm{v}}(\bm{x}_m^t)))$ denote the empirical source and the empirical target distributions in the embedding space that are built using the observed data points.  Similarly, let $\hat{\mu}_{\mathcal{P}}=\frac{1}{N_p}\sum_{q=1}^{N_p}\delta(\bm{z}_n^q)$ denote the empirical prototypical distribution which is built using the generated pseudo-dataset.  

  Finally, note that when fitting the GMM and when we generate the pseudo-dataset,   we only included those data points and pseudo-data points for which the model is confident about their predicted labels. For this reason, we can conclude that: $\tau =  \mathbb{E}_{\bm{z}\sim \hat{p_{J}(\bm{z})}}(\mathcal{L}(h(\bm{z}),h_{\hat{\bm{w}}_0}(\bm{z})) $.

\textbf{Theorem 1}: Consider that we generate a pseudo-dataset  using the prototypical distribution and update the model for sequential UDA using algorithm~\ref{SSUDAalgorithmSS}. Then, the following   holds:
\begin{equation}
\small
\begin{split}
e_{\mathcal{T}}\le & e_{\mathcal{S}} +W(\hat{\mu}_{\mathcal{S}},\hat{\mu}_{\mathcal{P}})+W(\hat{\mu}_{\mathcal{T}},\hat{\mu}_{\mathcal{P}})+(1-\tau)+e_{\mathcal{C'}}(\bm{w}^*)\\&+\sqrt{\big(2\log(\frac{1}{\xi})/\zeta\big)}\big(\sqrt{\frac{1}{N}}+\sqrt{\frac{1}{M }}+2\sqrt{\frac{1}{N_p }}\big),
\end{split}
\label{theroemforPLnipssemantic}
\end{equation}    
where $W(\cdot,\cdot)$ denotes the WD distance and   $\xi$ is a constant which depends on the loss function $\mathcal{L}(\cdot)$.

\textbf{Proof:}    the complete  proof is included in the Appendix.

  Theorem~1  justifies effectiveness of our algorithm. We observe that 
  MAS$^3$ algorithm minimizes the upperbound expressed in Eq.~\eqref{theroemforPLnipssemantic}. The source expected risk is minimized through the initial training on the source domain. The second term in Eq.~\eqref{theroemforPLnipssemantic} is minimized because we deliberately fit a GMM distribution on the source domain distribution in the embedding space. Note that minimizing this term is conditionally possible when the source domain distribution can be approximated well with  a GMM distribution. However, similar constraint exist for all  the parametric methods in statistics. Additionally, since we use a softmax in the last layer, this would likely to happen because the classes should become separable for a generalizable model to be trained. The third term in the Eq.~\eqref{theroemforPLnipssemantic} upperbound is minimized  as the second term in Eq.~\eqref{eq:mainPrMatchSS}. The fourth term is a constant term depending on the threshold we use and can be small if $\tau\approx 1$. Note that when selecting the source distribution samples for fitting the Gaussian distribution, if we set $\tau$ too close to 1, we may not have sufficient samples for accurate estimation of GMM and hence the second term may increase. Hence, there is a trade-off between minimizing the second and the fourth term in Eq.~\eqref{theroemforPLnipssemantic}.
  The term $e_{C'}(\bm{w}^*)$ will be small if the domains are related, i.e., share the same classes and the base model can generalize well in both domains, when trained in the presence of sufficient labeled data from both domains. In other words, aligning the distributions in the embedding must be a possibility for our algorithm to work. This is a condition for all UDA algorithms to work. Finally, the last term in Eq.~\eqref{theroemforPLnipssemantic} is a constant term similar to most PAC-learnability bounds and can be negligible  if       sufficiently large  source and target datasets are accessible and we generate a large pseudo-dataset.

\section{Experimental Validation}
We validate our algorithm using two benchmark domain adaptation tasks and compare it against existing algorithms.
\subsection{Experimental setup}

\textbf{Datasets and evaluation metrics:} We validate MAS$^3$ on the standard GTA5~\cite{richter2016playing}$\rightarrow$Cityscapes~\cite{cordts2016cityscapes} and the SYNTHIA~\cite{ros2016synthia}$\rightarrow$Cityscapes benchmark UDA tasks for semantic segmentation. 

\textbf{GTA5} consists of 24,966    $1914\times1052$ image instances. 

\textbf{SYNTHIA} consists of 9,400   $1280\times760$ image instances. 

\textbf{Cityscapes} is a real-world dataset consisting of a training set with 2,957  instances and a validation set, used as testing set, with 500   instances of images with size $2040\times 1016$.

The GTA5 and SYNTHIA datasets are used as source domains. After training the model, we adapt it to generalize on the Cityscapes dataset as the target domain. We resize all images  to $1024\times512$ size to use a shared cross-domain encoder. Implementation details  are included in the Appendix.

\textbf{Evaluation}: Following the   literature,  we report the results on the Cityscapes validation set and use the category-wise and the mean intersection over union (IoU) to measure segmentation performance~\cite{hoffman2016fcns}. Note  that while  GTA5  has the same 19 category annotations as Cityscapes, SYNTHIA has 16 common category annotations.  For this reason and following the literature,  we report the results on the shared cross-domain categories  for each task.  

\textbf{Comparison with the State-of-the-art Methods:} To the best of our knowledge, there is no prior source-free model adaptation algorithm for performance comparison. For this reason, we compare MAS$^3$ against UDA algorithms based on joint training due to proximity of these works to our learning setting.
In our comparison, we have included both pioneer and recent UDA image segmentation method to be representative of the literature. We have compared our performance against the adversarial learning-based UDA methods:  GIO-Ada~\cite{chen2018learning}, ADVENT~\cite{vu2018advent}, AdaSegNet~\cite{tsai2018learning}, TGCF-DA+SE \cite{choi2019selfensembling},  PCEDA~\cite{yang2020phase}, and CyCADA \cite{hoffman2017cycada}. We have also included  methods that are based on direct distributional matching which are more similar to MAS$^3$: FCNs in the Wild~\cite{hoffman2016fcns},  CDA~\cite{zhang2017curriculum}, DCAN~\cite{wu2018dcan},  SWD~\cite{lee2019sliced}, Cross-City~\cite{chen2017discrimination}.

\begin{table*}[!htb]
  \begin{adjustbox}{center}
  \scalebox{.85}{
      \begin{tabular} { @{} cr*{16}c   }
      \hline
       Method &Adv. &\rot{road} &\rot{sidewalk} &\rot{building} &\rot{traffic light} &\rot{traffic sign} &\rot{vegetation} &\rot{sky} &\rot{person} &\rot{rider}   &\rot{car} &\rot{bus} &\rot{motorcycle} &\rot{bicycle} &mIoU \\
       \hline
        Source Only (VGG16)  &N &6.4 &17.7 &29.7 &0.0 &7.2 &30.3& 66.8& 51.1& 1.5& 47.3& 3.9& 0.1& 0.0 &20.2\\
        FCNs in the Wild \cite{hoffman2016fcns} &N &11.5 &19.6 &30.8 &0.1 &11.7 &42.3 &68.7 &51.2 &3.8 &54.0 &3.2 &0.2 &0.6 &22.9 \\
        CDA \cite{zhang2017curriculum} &N &65.2 &26.1 &74.9 &3.7 &3.0 &76.1 &70.6 &47.1 &8.2 &43.2 &20.7 &0.7 &13.1 &34.8 \\
        DCAN \cite{wu2018dcan} &N &9.9 &30.4 &70.8  &6.70 &23.0 &76.9 &73.9 &41.9 &16.7 &61.7  &11.5 &10.3 &38.6 &36.4 \\
        SWD \cite{lee2019sliced} &N &83.3 &35.4 &82.1 &12.2 &12.6 &83.8 &76.5 &47.4 &12.0 &71.5 &17.9 &1.6 &29.7 &43.5 \\
        
        Cross-City \cite{chen2017discrimination} &Y &62.7 &25.6 &78.3 &1.2 &5.4 &81.3 &81.0 &37.4 &6.4 &63.5 &16.1 &1.2 &4.6 &35.7 \\
        GIO-Ada \cite{chen2018learning} &Y &78.3 &29.2 &76.9 &10.8 &17.2 &81.7 &81.9 &45.8 &15.4 &68.0 &15.9 &7.5 &30.4 &43.0 \\
        ADVENT \cite{vu2018advent} &Y &67.9 &29.4 &71.9 &0.6 &2.6 &74.9 &74.9 &35.4 &9.6 &67.8 &21.4 &4.1 &15.5 &36.6 \\
        AdaSegNet \cite{tsai2018learning} &Y &78.9 &29.2 &75.5 &0.1 &4.8 &72.6 &76.7 &43.4 &8.8 &71.1 &16.0 &3.6 &8.4 &37.6 \\
        TGCF-DA+SE \cite{choi2019selfensembling} &Y &90.1   &48.6 &80.7 &3.2 &14.3 &82.1 &78.4 &54.4 &16.4 &82.5 &12.3 &1.7 &21.8 &46.6 \\
        PCEDA \cite{yang2020phase} &Y &79.7 &35.2 &78.7 &10.0 &28.9 &79.6 &81.2 &51.2 &25.1 &72.2 &24.1  &16.7 &50.4 &48.7 \\
  \hline    
        MAS$^3$ (Ours)   &N &75.1 &49.6 &70.9 &14.1 &25.3 &72.7 &76.7 &48.5 &19.9 &65.3 &17.6 &6.8 &39.0 &44.7 \\
      \hline
      \end{tabular}
  }
  \end{adjustbox}
  \caption{Model adaptation comparison results  for the SYNTHIA$\rightarrow$Cityscapes task on 13 commonly used classes. The first row presents the source-trained model performance prior to adaptation to demonstrate the effect of knowledge transfer from the source domain.}
  \label{table:synthia}
\end{table*}

\subsection{Results}

\textbf{Quantitative  performance comparison:}

\textbf{SYNTHIA$\rightarrow$Cityscapes:} We report the   quantitative results in table \ref{table:synthia}. We note that despite addressing a more challenge learning setting, MAS$^3$ outperforms most of the UDA methods. Recently developed UDA methods based on adversarial learning outperform our method but we note that these methods benefit from a secondary type of regularization in addition to probability matching. Overall, MAS$^3$ performs reasonably well even compared with these UDA methods that need source samples. Additionally,  MAS$^3$ has the best performance for some important categories, e.g., traffic light.

\textbf{GTA5$\rightarrow$Cityscapes:} Quantitative results for this task are reported in Table~\ref{table:gta5}. We observe a more competitive performance for this task but  the performance comparison trend is similar.   These results demonstrate that although  the motivation in this work is source-free model adaptation, MAS$^3$ can also be used as a   joint-training UDA algorithm.   

\hspace{-4mm}\textbf{Qualitative  performance validation:}

\begin{figure*}[!htb]
    \centering
    \includegraphics[width=.92\textwidth]{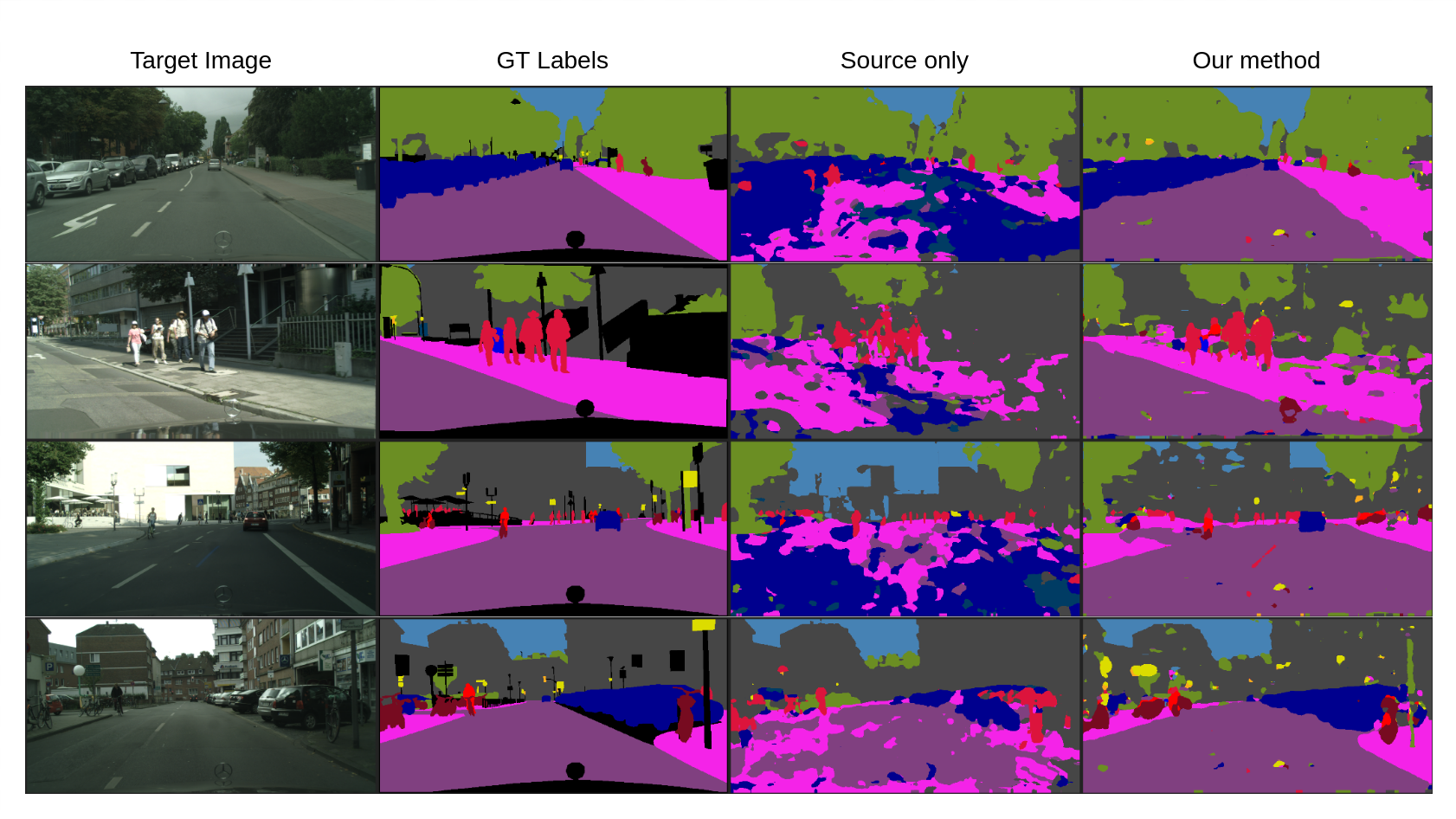}
    \caption{Qualitative performance: examples of the   segmented frames for SYNTHIA$\rightarrow$Cityscapes using the MAS$^3$ method. Left to right: real images, manually annotated images, source-trained model predictions, predictions based on our method.  }
    \label{figure:da}
\end{figure*}

In Figure~\ref{figure:da},  we have visualized   exemplar frames for   the Cityscapes dataset for the GTA5$\rightarrow$Cityscapes  task which are segmented using the model  prior and after adaptation along with the ground-truth (GT) manual annotation for each image. Visual observation demonstrates that our method is able to significantly improve image segmentation from the source-only segmentation to the post-adaptation segmentation, noticeably on sidewalk, road, and car semantic classes for the GTA5-trained model. 
 Examples of segmented frames     for the SYNTHIA$\rightarrow$Cityscapes  task are included in the Appendix and provide similar observation.

\hspace{-4mm}\textbf{Effect of alignment in the embedding space:}

\begin{figure}[!htb]
    \centering
    \subfloat[GMM samples]{
        \includegraphics[width=.14\textwidth]{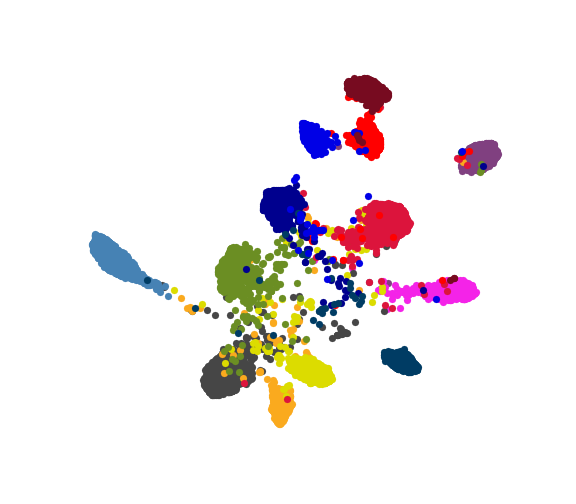}
        \label{figure:latent-features1}
    }
    \subfloat[Pre-adaptation]{
        \includegraphics[width=.14\textwidth]{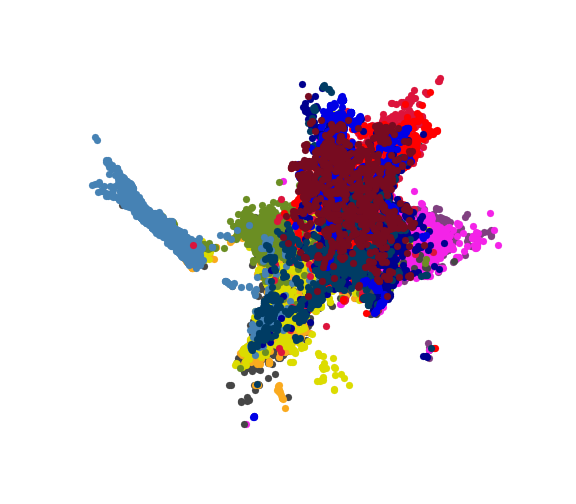}
        \label{figure:latent-features2}
    }
    \subfloat[Post-adaptation]{
        \includegraphics[width=.14\textwidth]{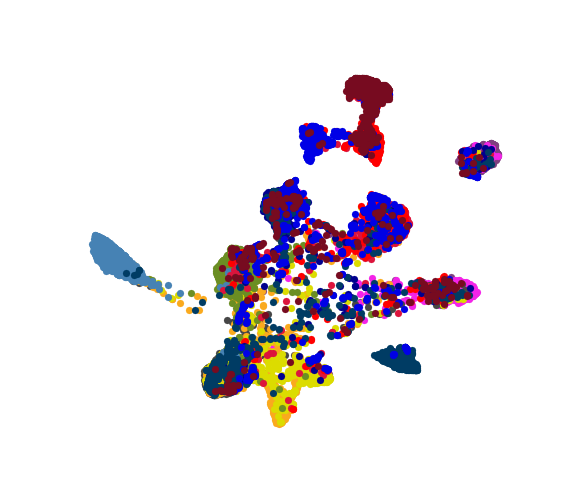}
        \label{figure:latent-features3}
    }
    \caption{Indirect distribution matching in the embedding space: (a) drawn samples  from the GMM trained on the SYNTHIA distribution, (b) representations of the Cityscapes validation samples prior to model adaptation (c) representation of the Cityscapes validation samples after domain alignment.}
    \label{figure:latent-features}
\end{figure}

To demonstrate that our solution implements what we anticipated, we  have used UMAP~\cite{mcinnes2018umap} visualization tool to reduce  the dimension of the data representations  in the  embedding space to two for 2D visualization. Figure~\ref{figure:latent-features} represents the samples  of the prototypical distribution along with the target domain data prior and after adaptation  in the embedding space for the  GTA5$\rightarrow$Cityscapes   task. Each point in Figure~\ref{figure:latent-features} denotes a single data point and each color denotes a semantic class cluster.
Comparing Figure~\ref{figure:latent-features2} and Figure~\ref{figure:latent-features3} with Figure~\ref{figure:latent-features1}, we can see that the semantic classes in the target domain have become much more well-separated and more similar to the prototypical distribution after model adaptation. This means that domain discrepancy has been reduced using MAS$^3$ and the source and the target domain distributions are aligned indirectly as anticipated  using the intermediate prototypical distribution in the embedding space.

\subsection{Ablation study}
\begin{figure}[!htb]
    \centering
    \subfloat[$\tau=0$ \newline mIoU=$41.6$]{
        \includegraphics[width=.14\textwidth]{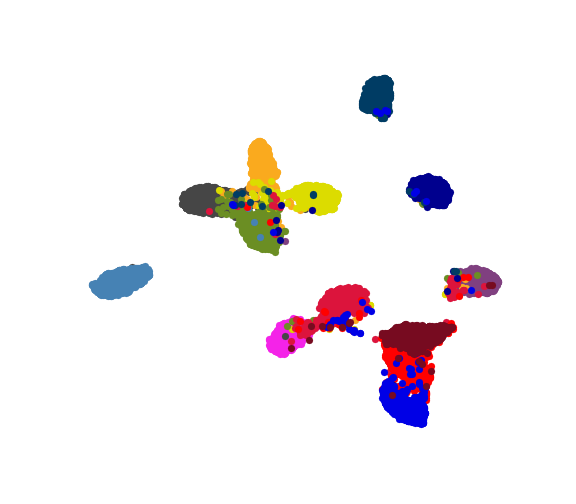}
    }
    \subfloat[$\tau=0.8$ \newline mIoU=$42.7$]{
        \includegraphics[width=.14\textwidth]{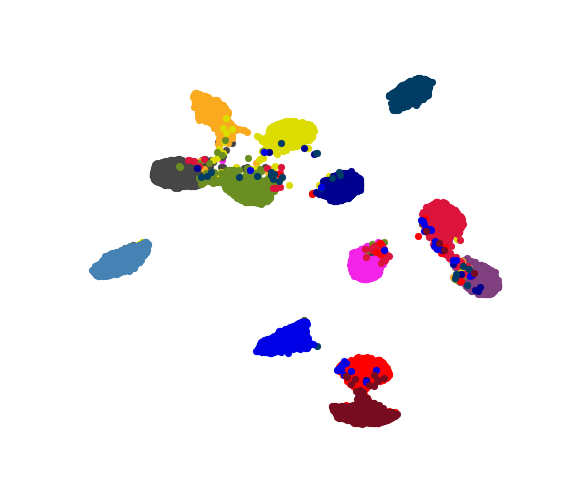}
    }
    \subfloat[$\tau=0.97$ \newline mIoU=$43.9$]{
        \includegraphics[width=.14\textwidth]{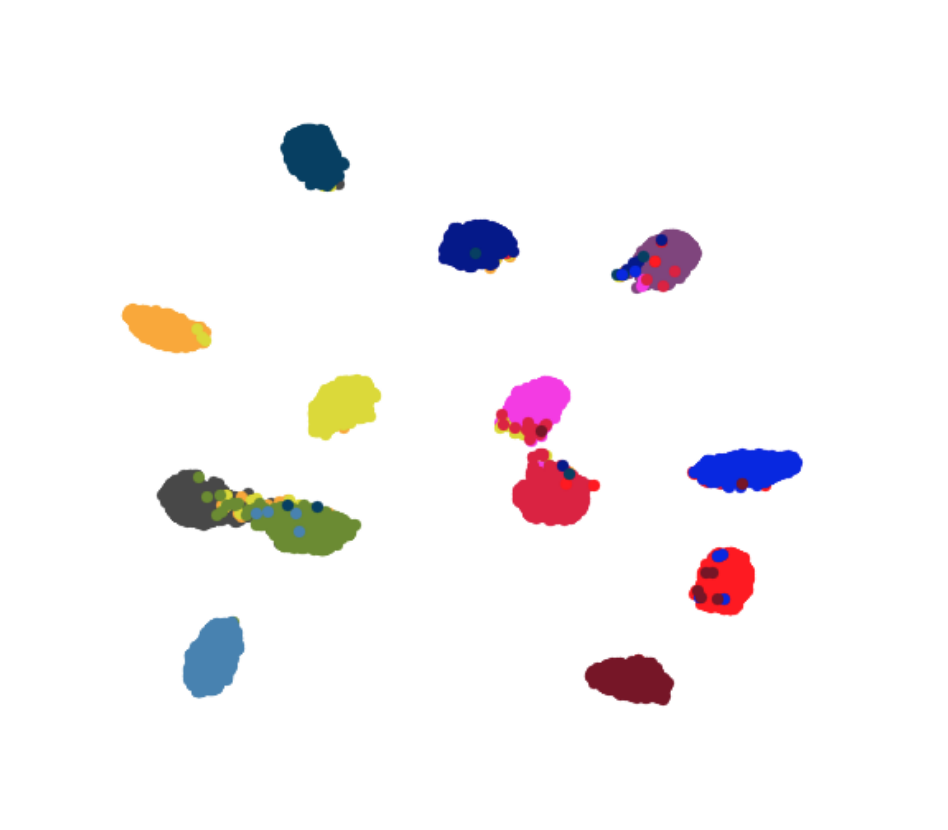}
    }
    \caption{Ablation experiment to study effect of $\tau$ on the GMM learnt in the embedding space: (a) all samples are used; adaptation mIoU=41.6, (b) a portion of samples is used; adaptation mIoU=42.7, (c) samples with high model-confidence are used; adaptation mIoU=43.9}
    \label{figure:rho-impact}
\end{figure}

A major advantage of our algorithm over methods based on adversarial learning is its simplicity in  depending on a few hyper-parameters. We note that the major algorithm-specific hyper-parameters are $\lambda$ and $\tau$. We observed in our experiments  that MAS$^3$ performance is stable with respect to the trade-off parameter $\lambda$ value. This is expected because in Eq.~\eqref{eq:mainPrMatchSS}, the $\mathcal{L}_{ce}$ loss term is small from the beginning due to prior training on the source domain. We  investigated the impact of the confidence hyper-parameter $\tau$ value.   Figure~\ref{figure:rho-impact} presents the fitted GMM on the source prototypical distribution for three different values of $\tau$. As it can be seen, when $\tau=0$, the fitted GMM clusters are cluttered. As we increase the threshold $\tau$ and use samples for which the classifier is confident, the fitted GMM represents well-separated semantic classes which increases knowledge transfer from the source domain. This experiments also empirically validates what we deduced  about importance of $\tau$ using Theorem~1.

\begin{table*}[!htb]
  \begin{adjustbox}{center}
  \scalebox{.76}{\footnotesize
      \begin{tabular} { @{} cr*{21}c   }
      \hline
       Method &Adv. &\rot{road} &\rot{sidewalk} &\rot{building} &\rot{wall} &\rot{fence} &\rot{pole} &\rot{traffic light} &\rot{traffic sign} &\rot{vegetation} &\rot{terrain} &\rot{sky} &\rot{person} &\rot{rider} &\rot{car} &\rot{truck} &\rot{bus} &\rot{train} &\rot{motorcycle} &\rot{bicycle} &mIoU \\
       \hline
        Source Only (VGG16) &N &25.9 &10.9 &50.5 &3.3 &12.2 &25.4 &28.6 &13.0 &78.3 &7.3 &63.9 &52.1 &7.9 &66.3 &5.2 &7.8 &0.9 &13.7 &0.7 &24.9 \\
        FCNs in the Wild \cite{hoffman2016fcns} &N &70.4 &32.4 &62.1 &14.9 &5.4 &10.9 &14.2 &2.7 &79.2 &21.3 &64.6 &44.1 &4.2 &70.4 &8.0 &7.3 &0.0 &3.5 &0.0 &27.1 \\
        CDA \cite{zhang2017curriculum} &N &74.9 &22.0 &71.7 &6.0 &11.9 &8.4 &16.3 &11.1 &75.7 &13.3 &66.5 &38.0 &9.3 &55.2 &18.8 &18.9 &0.0 &16.8 &14.6 &28.9 \\
        DCAN \cite{wu2018dcan} &N &82.3 &26.7 &77.4 &23.7 &20.5 &20.4 &30.3 &15.9 &80.9 &25.4 &69.5 &52.6 &11.1 &79.6 &24.9 &21.2 &1.30 &17.0 &6.70 &36.2 \\
        SWD \cite{lee2019sliced} &N &91.0 &35.7 &78.0 &21.6 &21.7 &31.8 &30.2 &25.2 &80.2 &23.9 &74.1 &53.1 &15.8 &79.3 &22.1 &26.5 &1.5 &17.2 &30.4 &39.9 \\
        
        CyCADA \cite{hoffman2017cycada} &Y &85.2 &37.2 &76.5 &21.8 &15.0 &23.8 &22.9 &21.5 &80.5 &31.3 &60.7 &50.5 &9.0 &76.9 &17.1 &28.2 &4.5 &9.8 &0.0 &35.4 \\
        ADVENT \cite{vu2018advent} &Y &86.9 &28.7 &78.7 &28.5 &25.2 &17.1 &20.3 &10.9 &80.0 &26.4 &70.2 &47.1 &8.4 &81.5 &26.0 &17.2 &18.9 &11.7 &1.6 &36.1 \\
        AdaSegNet \cite{tsai2018learning} &Y &86.5 &36.0 &79.9 &23.4 &23.3 &23.9 &35.2 &14.8 &83.4 &33.3 &75.6 &58.5 &27.6 &73.7 &32.5 &35.4 &3.9 &30.1 &28.1 &42.4 \\
        TGCF-DA+SE \cite{choi2019selfensembling} &Y &90.2 &51.5 &81.1 &15.0 &10.7 &37.5 &35.2 &28.9 &84.1   &32.7 &75.9 &62.7 &19.9 &82.6 &22.9 &28.3 &0.0 &23.0 &25.4 &42.5 \\
        PCEDA \cite{yang2020phase} &Y &90.2 &44.7 &82.0 &28.4 &28.4 &24.4 &33.7 &35.6 &83.7 &40.5 &75.1 &54.4 &28.2 &80.3 &23.8 &39.4 &0.0 &22.8 &30.8 &44.6 \\
        \hline
        MAS$^3$ (Ours)   &N &75.5 &53.7 &72.2 &20.5 &24.1 &30.5 &28.7 &37.8 &79.6 &36.9 &78.7 &49.6 &16.5 &77.4 &26.0 &42.6 &18.8 &15.3 &49.9 &43.9\\
      \hline
      \end{tabular}
  }
  \end{adjustbox}
  
  \caption{Domain adaptation results for different methods for the GTA5$\rightarrow$Cityscapes task.}
  \label{table:gta5}
\end{table*}

\section{Conclusions}
 We developed an algorithm for adapting a image segmentation model to generalize in new domains after training using solely unlabeled data. Our algorithm is based on using an intermediate multi-modal prototypical distribution to minimize the the distributional cross-domain discrepancy in a shared embedding space. We estimate the prototypical distribution as a parametric GMM distribution. Experiments on benchmark tasks demonstrate our algorithm is effective and leads to competitive performance, even when compared to UDA algorithms that are based on joint-domain model training.   


\clearpage

\bibliographystyle{aaai} 
\bibliography{ref}

\section{Appendix}

\subsection{Proof of Theorem~1}
 Our proof is based on   the following theorem by Redko et al.~\citeyear{redko2017theoretical} which relates the performance of a trained model in a target domain to its performance to the source domain.

\textbf{Theorem~2 (Redko et al.~\cite{redko2017theoretical})}: Under the assumptions described in our framework, assume that a model is trained on the source domain, then for any $d'>d$ and $\zeta<\sqrt{2}$, there exists a constant number $N_0$ depending on $d'$ such that for any  $\xi>0$ and $\min(N,M)\ge \max (\xi^{-(d'+2),1})$ with probability at least $1-\xi$, the following holds:
\begin{equation}
\begin{split}
e_{\mathcal{T}}\le & e_{\mathcal{S}} +W(\hat{\mu}_{\mathcal{T}},\hat{\mu}_{\mathcal{S}})+e_{\mathcal{C}}(\bm{w}^*)+ \\& \sqrt{\big(2\log(\frac{1}{\xi})/\zeta\big)}\big(\sqrt{\frac{1}{N}}+\sqrt{\frac{1}{M}}\big).
\end{split}
\label{segmenteq:theroemfromcourty}
\end{equation}    
 
Theorem~2 provides an upperbound for the performance of the  model on the target domain in terms of the source true expected error and the distance between the source and the target domain distributions when measured in WD distance. We use Theorem~2 to deduce Theorem~1. Following Redko et al.~\citeyear{redko2017theoretical}, our analysis has been preformed for the case of binary classifier but it can be conveniently extended.

\textbf{Theorem~1 }:  Consider that we generate a pseudo-dataset  using the prototypical distribution and the confidence parameter $\tau$. If we adapt the model  using MAS$^3$ algorithm,  the following   holds:
\begin{equation}
\small
\begin{split}
e_{\mathcal{T}}\le & e_{\mathcal{S}} +W(\hat{\mu}_{\mathcal{S}},\hat{\mu}_{\mathcal{P}})+W(\hat{\mu}_{\mathcal{T}},\hat{\mu}_{\mathcal{P}})+(1-\tau)+e_{\mathcal{C'}}(\bm{w}^*)+\\&\sqrt{\big(2\log(\frac{1}{\xi})/\zeta\big)}\big(\sqrt{\frac{1}{N}}+\sqrt{\frac{1}{M }}+2\sqrt{\frac{1}{N_p }}\big),
\end{split}
\label{segmenteq:theroemforPLnips}
\end{equation}    
where    $\xi$ is a constant which depends on $\mathcal{L}(\cdot)$ and  $e_{C'}(\bm{w}^*)$  denotes the expected risk of the optimally   trained model which is trained jointly on both domains when labeled data is accessible in both domains.

\textbf{Proof:} Note that we use the confidence parameter $\tau$ to ensure that we only select the pseudo-data points for which the model is confident. Hence, the probability of  predicting incorrect  labels for the pseudo-data points by the classifier model is $1-\tau$.  We   define the following difference for a given pseudo-data point:
\begin{equation}
\begin{split}
  |\mathcal{L}(h_{\bm{w}_0}(\bm{z}^p_i),\bm{y}^p_i)- \mathcal{L}(h_{\bm{w}_0}(\bm{z}^p_i),\hat{\bm{y}}_i^{p})|= \begin{cases}
    0, & \text{if $\bm{y}^p_i=\hat{\bm{y}}_i^{p}$}.\\
    1, & \text{otherwise}.
  \end{cases}
\end{split}
\label{segmenteq:theroemforPLproof}
\end{equation}    
 Now using Jensen's inequality and by applying the expectation operator with respect to the target domain distribution in the embedding space, i.e., $\psi(\phi(P_{\mathcal{T}}(\bm{X}^t)))$, on both sides of above error function, we can deduce:
\begin{equation}
\begin{split}
&|e_{\mathcal{P}}-e_{\mathcal{T}}|\le\\&\mathbb{E}_{\bm{z}^p_i\sim \psi(\phi(P_{\mathcal{T}}))}\big(|\mathcal{L}(h_{\bm{w}_0}(\bm{z}^p_i),\bm{y}^p_i)- \mathcal{L}(h_{\bm{w}_0}(\bm{z}^p_i),\hat{\bm{y}}_i^{p})|\big)\le  \\&
(1-\tau).
\end{split}
\label{segmenteq:theroemforPLproofexpectation}
\end{equation}    
Using Eq.~\eqref{segmenteq:theroemforPLproofexpectation} we can deduce the following:
\begin{equation}
\begin{split}
&e_{\mathcal{S}}+e_{\mathcal{T}}=e_{\mathcal{S}}+e_{\mathcal{T}}+e_{\mathcal{P}}-e_{\mathcal{P}}\le  
e_{\mathcal{S}}+e_{\mathcal{P}}+|e_{\mathcal{T}}-e_{\mathcal{P}}|\le\\&  
e_{\mathcal{S}}+e_{\mathcal{P}}+(1-\tau).
\end{split}
\label{segmenteq:theroemforPLprooftrangleinq}
\end{equation}    
  Eq.~\eqref{segmenteq:theroemforPLprooftrangleinq} is valid for all $\bm{w}$, so by taking infimum on both sides of Eq.~\eqref{segmenteq:theroemforPLprooftrangleinq} and using the definition of the joint optimal model, we deduce the following:
\begin{equation}
\begin{split}
e_C(\bm{w}^*)\le e_{\mathcal{C'}}(\bm{w})+(1-\tau).
\end{split}
\label{segmenteq:theroemforPLprooftartplerror}
\end{equation}    
Now consider
Theorem~2 for the source and target domains and  apply Eq.~\eqref{segmenteq:theroemforPLprooftartplerror}  on Eq.\eqref{segmenteq:theroemfromcourty}, then we conclude:
\begin{equation}
\begin{split}
e_{\mathcal{T}}\le & e_{\mathcal{S}} +W(\hat{\mu}_{\mathcal{T}},\hat{\mu}_{\mathcal{S}})+e_{\mathcal{C'}}(\bm{w}^*)+ (1-\tau) \\&+ \sqrt{\big(2\log(\frac{1}{\xi})/\zeta\big)}\big(\sqrt{\frac{1}{N}}+\sqrt{\frac{1}{M}}\big),
\end{split}
\label{segmenteq:theroemfromcourty1}
\end{equation}    
where $e_{\mathcal{C'}}$ denotes the joint optimal model true error for the source and the pseudo-dataset.

Now we apply the triangular inequality twice in Eq.~\eqref{segmenteq:theroemfromcourty1} on considering that the WD is a metric, we  deduce:
\begin{equation}
\begin{split}
& W(\hat{\mu}_{\mathcal{T}},\hat{\mu}_{\mathcal{S}})\le  W(\hat{\mu}_{\mathcal{T}},\mu_{\mathcal{P}})+W(\hat{\mu}_{\mathcal{S}},\mu_{\mathcal{P}})  \le\\& W(\hat{\mu}_{\mathcal{T}},\hat{\mu}_{\mathcal{P}})+W(\hat{\mu}_{\mathcal{S}},\hat{\mu}_{\mathcal{P}})+2W(\hat{\mu}_{\mathcal{P}},\mu_{\mathcal{P}}) .
\end{split}
\label{segmenteq:theroemfromcourty2}
\end{equation}

 We then use Theorem 1.1 in the work by Bolley et al.~\citeyear{bolley2007quantitative} and  simplify the term $W(\hat{\mu}_{\mathcal{P}},\mu_{\mathcal{P}})$.

  \textbf{Theorem~3} (Theorem 1.1 by Bolley et al.~\cite{bolley2007quantitative}): consider that $p(\cdot) \in\mathcal{P}(\mathcal{Z})$ and $\int_{\mathcal{Z}} \exp{(\alpha \|\bm{x}\|^2_2)}dp(\bm{x})<\infty$ for some $\alpha>0$. Let $\hat{p}(\bm{x})=\frac{1}{N}\sum_i\delta(\bm{x}_i)$ denote the empirical distribution that is built from the samples $\{\bm{x}_i\}_{i=1}^N$ that are drawn i.i.d from $\bm{x}_i\sim p(\bm{x})$. Then for any $d'>d$ and $\xi<\sqrt{2}$, there exists $N_0$ such that for any $\epsilon>0$ and $N\ge N_o\max(1,\epsilon^{-(d'+2)})$, we have:
 \begin{equation}
\begin{split}
P(W(p,\hat{p})>\epsilon)\le \exp(-\frac{-\xi}{2}N\epsilon^2)
\end{split}
\label{segmenteq:mainSuppICML3}
\end{equation}  
 This relation measures the distance between the estimated empirical distribution and the true distribution when measured by the WD distance.

 We can use both   Eq.~\eqref{segmenteq:theroemfromcourty2} and Eq.~\eqref{segmenteq:mainSuppICML3} in Eq.~\eqref{segmenteq:theroemfromcourty1} and  conclude Theorem~2 as stated:
\begin{equation}
\small
\begin{split}
e_{\mathcal{T}}\le & e_{\mathcal{S}} +W(\hat{\mu}_{\mathcal{S}},\hat{\mu}_{\mathcal{P}})+W(\hat{\mu}_{\mathcal{T}},\hat{\mu}_{\mathcal{P}})+(1-\tau)+e_{\mathcal{C'}}(\bm{w}^*)\\&+\sqrt{\big(2\log(\frac{1}{\xi})/\zeta\big)}\big(\sqrt{\frac{1}{N}}+\sqrt{\frac{1}{M }}+2\sqrt{\frac{1}{N_p }}\big),
\end{split}
\label{segmenteq:theroemforPLnips55}
\end{equation}

\subsection{Details of Experimental Implementation}

Following standard approaches from literature ~\cite{tsai2018learning, lee2019sliced}, we have used the GTA5 and the SYNTHIA-RAND-CITYSCAPES datasets as source domains and the Cityscapes dataset as target domain. For our feature extractor we have used DeepLabV3~\cite{chen2017rethinking} with a VGG16 ~\cite{simonyan2014very} as a backbone. 
Training the VGG16-based feature extractor was done for both source datasets using an Adam optimizier, with learning rate $lr = 10^{-4}$ and epsilon $\epsilon =\{10^{-1}, 10^{-4}$ and $10^{-8}\}$. We used a training schedule so that the model is trained with each epsilon (from highest to lowest) for at least $50,000$ epochs. For source training, we used a batch size of size $4$.

When learning the GMM from the source data representations embedding, we tuned $\tau$  for better adaptation results. Quantitative results reported in this work are obtained by using $\tau=0.97$. As seen in Figure~4, lowering $\tau$ may negatively impact performance but values $\tau=+0.95$ works reasonably well. 

For model adaptation we  used the Adam optimizer, with learning rate $lr = 10^{-4}$ and epsilon $\epsilon = 10^{-1}$. We set the SWD loss regularization parameter to $0.5$ and used $100$ random projections. We use a batch size of $2$ for   images and a sample size from the GMM which is proportional to the per-batch label distribution. We use this as a surrogate for the target label distribution due to GPU memory constraints.

Experiments were conducted on an Nvidia Titan Xp GPU. We will provide experimental code at a publicly accessible domain.   

\subsection{Additional Results}

\subsubsection{Qualitative Results}

\begin{figure*}[ht]
    \centering
    \includegraphics[width=.92\textwidth]{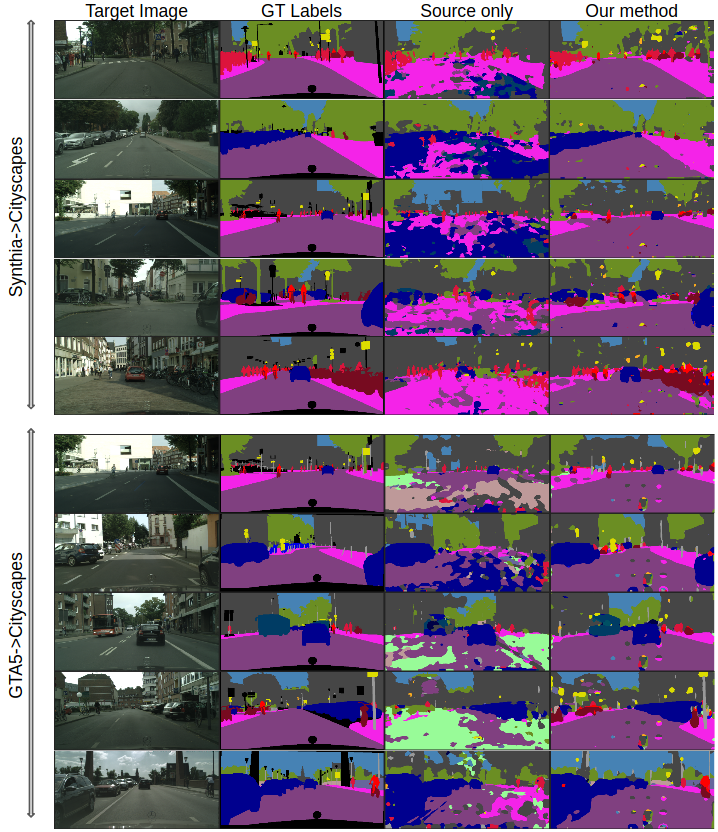}
    \caption{Qualitative performance: examples of the segmented frames for SYNTHIA$\rightarrow$Cityscapes and GTA5$\rightarrow$Cityscapes using the MAS$^3$ method. From left to right column: real images, manually annotated images, source-trained model predictions, predictions based on our method.  }
    \label{figure:da-extended}
\end{figure*}

We provide an extension to the semantic map visualizations in the main body of the paper, and include more image instances for both SYNTHIA$\rightarrow$Cityscapes and GTA5$\rightarrow$Cityscapes in Figure \ref{figure:da-extended}.

\end{document}